\documentclass[10pt, twocolumn, letterpaper]{article}

\usepackage{cvpr}
\usepackage{times}
\usepackage{epsfig}
\usepackage{graphicx}
\usepackage{amsmath}
\usepackage{amssymb}
\usepackage{balance}

\usepackage{hyperref}

\cvprfinalcopy

\begin{document}

\title{Efficient Urdu Caption Generation using Attention based LSTM}

\author{
Inaam Ilahi, H.M. Abdullah Zia, M. Ahtazaz Ahsan, M.
Rauf Tabassam, Armaghan Ahmad,\\
Information Technology University (ITU),\\
Lahore, Pakistan
}

\maketitle

\begin{abstract}
Recent advancements in deep learning have created many opportunities to solve real-world problems that remained unsolved for more than a decade. Automatic caption generation is a major research field, and the research community has done a lot of work on it in most common languages like English. Urdu is the national language of Pakistan and also much spoken and understood in the sub-continent region of Pakistan-India, and yet no work has been done for Urdu language caption generation. Our research aims to fill this gap by developing an attention-based deep learning model using techniques of sequence modeling specialized for the Urdu language. We have prepared a dataset in the Urdu language by translating a subset of the "Flickr8k" dataset containing 700 'man' images. We evaluate our proposed technique on this dataset and show that it can achieve a BLEU score of 0.83 in the Urdu language. We improve on the previous state-of-the-art by using better CNN architectures and optimization techniques. Furthermore, we provide a discussion on how the generated captions can be made correct grammar-wise. \footnote{\textit{\textbf{This is a project report for the Deep Learning Course (Spring 2020) being taught at Information Technology University, Lahore, Pakistan\\Corresponding email: inaam.ilahi@itu.edu.pk}}}

\end{abstract}

\section{Introduction}

With the evolution of Artificial Intelligence (AI) and deep learning, research community is now moving to those problems which can help us in real world gains. From the image classification tasks to the object detection, deep learning has played a vital role. On the other side, Natural language processing (NLP) has also created wide range of applications starting from simple text classification to fully automated chat-bots in native languages.

Caption generation is a rising research field which combines computer vision with NLP. It has a number of applications: $(i)$ transcribing the scenes for blind people, $(ii)$ classifying videos \& images based on different scenarios, $(iii)$ image based search engines for an optimized search, $(iv)$ visual question answering , and $(v)$ context understanding. Although, huge amount of research work has been done in this field for languages like English, French etc but there is no current research or publication that focuses on the Urdu language caption generation. Urdu language is spoken by more than 100 million people of Pakistan and India. This serves as a motivation to our problem of generating Urdu captions to remove the language barriers. This will help the natives in understanding visuals in many real world applications.

\begin{figure}[!t]
  \centering
  \includegraphics[width=0.55\linewidth]{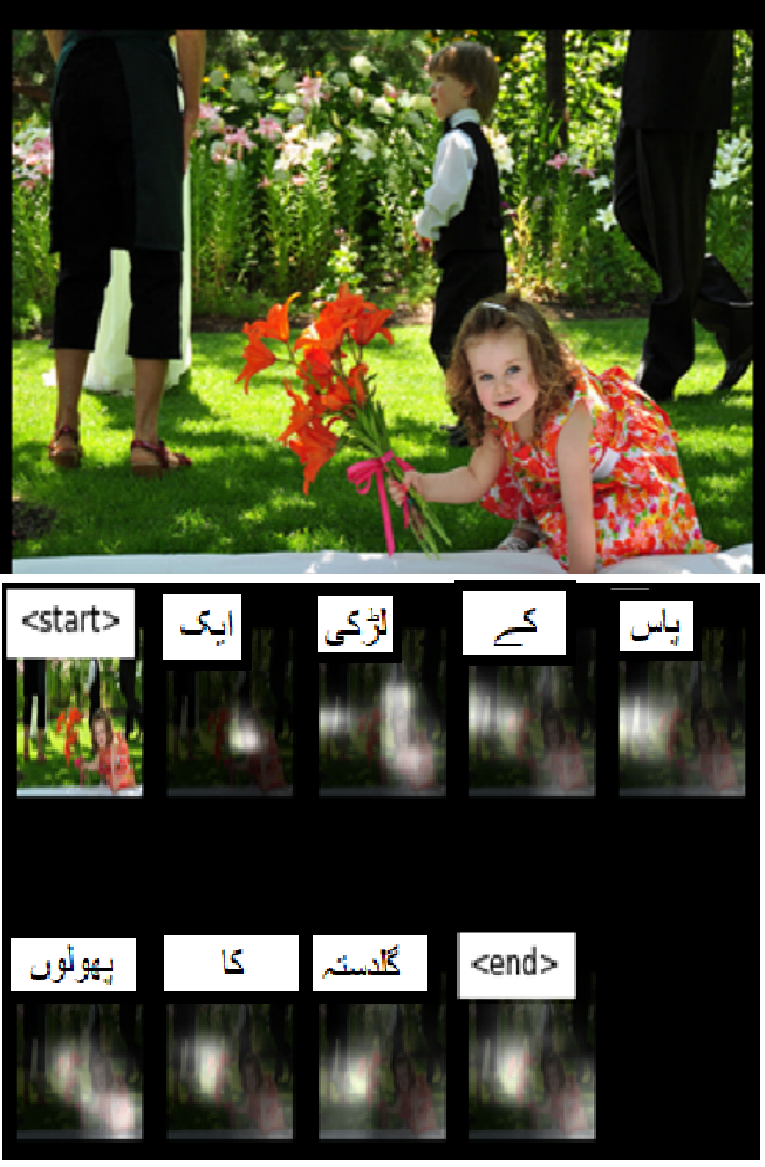}
  \caption{Result of soft attention of our model. Figure shows each word against the place of image where the attention-mechanism proposed the model to pay attention to.}
  \label{fig:soft}
\end{figure}

There are several challenges associated with Urdu language caption generation. Some of them are:
\begin{itemize}
    \item Urdu has its own set of syntactical, structural, and grammatical rules which must be followed to generate a proper caption.
    \item There are many ways to write a single word by using different language symbols.
\end{itemize}

Several image-based datasets are available containing multiple captions against each image in English language: $(i)$ Flickr8k \cite{hodosh2013framing}, $(ii)$ Flickr30k \cite{flickr30k}, and $(iii)$ COCO \cite{lin2014microsoft}. \cite{hodosh2013framing} contains 8k images with 5 English captions against each image. \cite{flickr30k} has 30k images with 5 English captions against each image. \cite{lin2014microsoft} contains around 200k labelled images. Owing to being short on time, we chose to work with Flickr8k \cite{hodosh2013framing} to work on. Different practices were followed when translating the captions to Urdu which are detailed in the following sections.

The major contributions of our paper are:
\begin{itemize}
    \item We prepare a dataset of Urdu language captions.
    \item We evaluate previous state-of-the-art techniques proposed for the task of caption generation.
    \item We enhance their performance by combining them with newly proposed techniques.
    \item We do extensive testing and provide a detailed discussion of the results.
\end{itemize}

The rest of the paper is organized as: Section \ref{sec:related work} provides a background of previously proposed techniques. Section \ref{sec:sol} we discuss our proposed mechanism. In Section \ref{sec:results}, we evaluate our approach and show the results. In Section \ref{sec:discussion}, we discuss the future directions in detail. The paper is concluded in Section \ref{sec:conclusion}. 

\section{Related Work}
\label{sec:related work}

In pre-deep learning era, people used template-based caption generation techniques. Templates are filled by detecting different objects, and attributes. Farhadi et al. \cite{farhadi2010every} proposed triplet of scene elements to fill the template. Li et al. \cite{li2011composing} used a method to extract the phrases related to detected objects, attributes and their relationships. Kulkarni et al. \cite{bybaby} proposed a conditional random field (CRF) to infer the objects, attributes, and prepositions before filling in the gaps. Although template based methods generate grammatically correct captions but they lack in providing flexibility to change the length of text. Retrieval based caption generation was introduced in \cite{kuznetsova2012collective}. Visually similar images and their captions are retrieved first from a large database and then captions for queried image are generated. This method produces correct and generalizable captions but it can not generate semantically correct and mage specific captions. 

Use of deep neural networks (DNNs) for caption generation was first proposed by Kiros \cite{kiros2014multimodal}. They proposed to use convolutional neural networks (CNNs) for extracting features from images to generate captions. They use a multi-modal space to represent images and text jointly for multi-modal representation learning and image caption generation. This model use high level features which provide more information, do not use fixed length templates, and use multi-modal neural language models for improving the results on language part. \cite{jozefowicz2016exploring} argue that neural language models can not handle large amount of data and to mitigate this problem, \cite{kiros2014unifying} proposed LSTM based image caption generation model that can handle the long dependencies in a sentence. Furthermore, they propose a neural language model for image caption generation: "structure-content neural language model (SC-NLM)".

LSTMS and RNNs removed the barriers in sequence to sequence sentence generation. Karpathy et al. \cite{karpathy2014deep} proposed a model that learns a joint embedding space for ranking and generation. Their model learns to score sentences and image similarity as a function of R-CNN object detection with outputs of a bidirectional recurrent neural networks (RNNs). This method works at a finer level and embeds fragments of images and sentences. This method breaks down the images into a number of objects and sentences into a dependency tree relations (DTR) \cite{de2006generating}. Afer that, it reasons about their latent and inter-modal alignment. It showed that the method achieves significant improvements in the retrieval task compared to other previous methods. Despite of the great performance, this method faces a few limitations in the generation of dependency trees as the model relations are not always appropriate.

Mao et al. \cite{mao2014deep} proposed multi-modal RNNs for image caption generation. Their proposed method uses 2 types of neural networks: a deep CNN for images and a deep RNN for sentences. These two networks interact with each other in a multi-modal layer to form the whole model. Both image and of sentences are given as input. It calculates the probability distribution to generate the next word of the caption. Vinyals et al. \cite{vinyals2015show} proposed a method called Neural Image Caption Generator. They propose to use a CNN as an encoder for feature extraction from images and a long short term memory (LSTM) for generating captions. The output of the last hidden layer of the encoder is used as the input to the LSTM. The LSTM is capable of keeping track of the objects that have been described by the text. The model is trained based on maximum likelihood estimation (MLE) methods and uses joint embedding to generate image captions.

Fang et al. \cite{fang2015captions} proposed a three-step pipeline for caption generation by incorporating object detection. Their model first learn detectors for several visual concepts based on a multi-instance learning framework. A language model trained on captions was then applied to the detector outputs, followed by re-scoring from a joint image-text embedding space.

Xu et al. \cite{xu2015show} proposed to use the output from the convolutional layers instead of last hidden FC layers of the CNN. They introduced an attention-based image captioning method for the first time. These methods can concentrate on the salient parts of the image and generate the corresponding words at the same time. Two types of attention mechanisms are proposed in this paper: stochastic hard attention and deterministic soft attention. It uses convolutional layers of CNN as encoder for extracting information of the salient objects from the image. This method does not use fully connected layers, so that it can focus only on salient objects from images. Attention mechanism helps to train model so that it can detect which word describes which part of an image. LSTM is used as decoder that generate captions.

We focus on generating captions on images in Urdu language and there is no publication available for the same. This makes our work unique. Our work is much related to the previous work done in English language proposed by Xu et al. \cite{xu2015show}. We train our caption generation model on new CNN architectures to extract more features from images to overcome the disadvantages of this model. We also use different optimization techniques in order to get better results. Furthermore, we replace LSTM in hard attention models with gated recurrent unit (GRU) \cite{cho2014learning}.

\section{Proposed Solution}
\label{sec:sol}

Our focus of work is on \textit{Whole Scene}-based captioning. As discussed in the introduction section, there is no dataset or a good translating API available which provides the captions in Urdu. So, we divide our project into three parts: the first part focuses on the development of a dataset containing Urdu captions, the second part aims at developing a deep learning model which is sufficiently able to generate understandable Urdu captions for a given image, and the last part is to critically test this model. Furthermore, we also try performing some architectural changes in best model for improved performance. These changes are discussed in the deep learning model subsection. 

\begin{figure*}[t]
  \centering
  \includegraphics[width=0.78\linewidth]{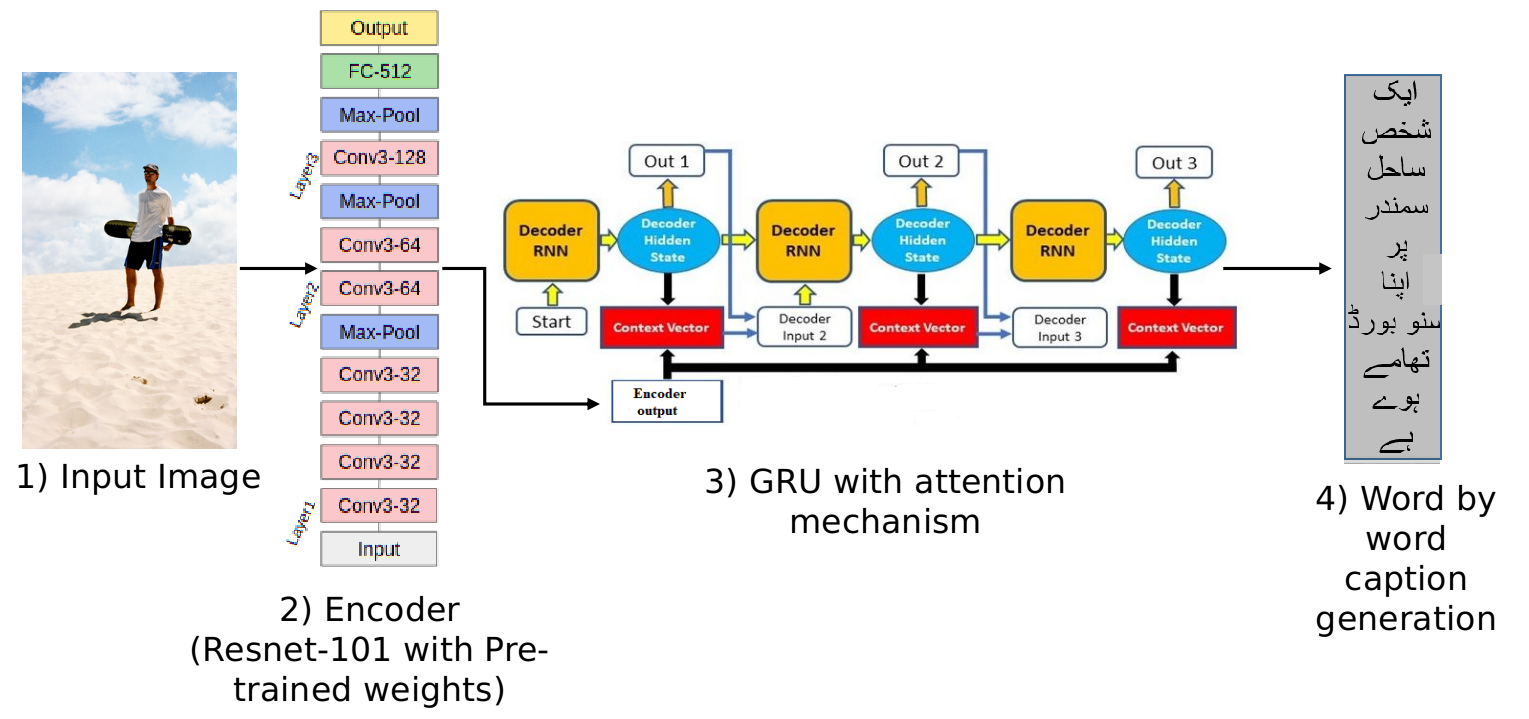}
  \caption{Figure showing our proposed architecture. It is inspired by the architecture proposed by Xu et al. \cite{xu2015show}. The input image is first passed into the encoder from which the output of the convolutional layers is passed into the GRU with attention mechanism. The proposed words are then joined to form a complete caption against the image. These captions are then passed to the grammar testing DMM.}
  \label{fig:architecture}
\end{figure*}

Our final proposed architecture has been shown in Figure \ref{fig:architecture}. The details of each part of the model are discussed in the following subsections.

\subsection{Dataset Generation}

We use the Flickr8k dataset to build our dataset. Following practices were followed in the creation of the dataset:
\begin{itemize}
    \item The captions were divided in the people of the group, i.e., all the captions were not translated by the same person. 
    \item All the captions of a certain image were not translated at the same time.
    \item In case of a discrepancy of English translations from the images, the Urdu captions were written from scratch on the base of the image.
    \item After successfully translating the captions by a person, they were reviewed by other people of the group.
\end{itemize}

This data was then preprocessed in order to be used as input to the DL model. This preprocessing involved: (i) removing the punctuation, and (ii) tokenizing the dataset. Tokenizing with no spaces present among the words is a difficult task in Urdu, as there are two types of: $(i)$ non-joiners and $(ii)$ joiners. The later change their shape depending on the part of the word they exist in. The former retain their shape despite of which part of the word they are present in. Even with joiners we can not be sure that the word is ending as they can also exist in the center of the word. There are several theorems available for tokenizing different languages but none provides any code that might be useful in our problem of Urdu tokenization. So, adding spaces between the words was kept in consideration while writing the captions. These tokens were then further processed and $<start>$ \& $<end>$ were added to each of the sentence.

\subsection{Deep Learning Model}
Our technique is hugely inspired by the technique of Xu et al. \cite{xu2015show}. We have improved upon their performance. Our model consists of three major parts: encoder, attention mechanism, and decoder. The input to the encoder is the images with the captions. The output of the encoder is passed to the attention mechanism and the LSTM that work in parallel to predict captions of the input image.

Encoder consists of the CNN which extracts features from image. The features are extracted from the last convolutional layer instead of the fully connected layers. The output of this CNN is fed to the attention mechanisms. The advantage of taking the output of the last convolutional layer is that we can extract salient features that might otherwise be lost in case of taking output from the fully connected layers. The latest CNN architectures including ResNet-101 \cite{he2016deep}, DenseNet-161 \cite{huang2017densely}, and InceptionV3 \cite{szegedy2016rethinking} were used as CNN. ResNet-101 \& InceptionV3 performed the best hence they were chosen for further testing. Using these architectures we were able to extract comparatively more features. We also tried different optimizers: namely, Momentum \cite{sutskever2013importance}, Adam \cite{kingma2014adam}, and RMSprop \cite{hinton2012neural}.

With the help of an attention mechanism, the model can learn to focus on the relevant part of the image. We used the attention made by Bahdanau at el. \cite{bahdanau2014neural}. This attention mechanism was termed as "Soft Attention" by Xu et al. \cite{xu2015show}. This attention is deterministic, i.e., it pays equal attention to all parts of the image. Image features and previous hidden state of decoder are passed to attention mechanism. The alignment score (AS) is first calculated by the attention mechanism. This score tells the decoder about how much attention decoder has to pay on a particular part of the image. The alignment score is calculated as:

\begin{equation}
AS = W_c \times tanh((W_D \times H_D) + (W_E \times X_E))
\label{eq:alsc}
\end{equation}

This score is then passed through a softmax in order to calculate the weighted attention score ($AS_W$) as shown in equation \ref{eq:sftmax}.

\begin{equation}
AS_W = softmax(AS)
\label{eq:sftmax}
\end{equation}

Then, context vector is generated by doing an element-wise multiplication of the attention weights with the encoder outputs (extracted features) as shown in equation \ref{eq:cv}. Due to the softmax, if the score of a specific input element is closer to 1, its effect and influence on the decoder output is amplified, whereas if the score is close to 0, its influence is drowned out and nullified.

\begin{equation}
Context\:Vector = AS_W \times X_E
\label{eq:cv}    
\end{equation}

The context vector is further passed to the decoder where it is concatenated with current input of decoder. The concatenated output is then passed to GRU to generate next word and this process repeats until last word of the caption is generated. These generated words are then concatenated to generate a complete sentence.

The decoder consists of a GRU \cite{cho2014learning} to generate captions. GRU is similar to LSTM but has fewer parameters. GRU also does not have an output gate. These features make GRU faster, computationally inexpensive, and memory efficient. GRU generates one word at each time step. This generated word is conditioned on the previous hidden state of GRU, previously generated word, and context vector. Chung et al. \cite{chung2014empirical} proved the GRU show more accurate results on small datasets than the LSTMs. This makes GRU the best choice for our problem.

\section{Experimental Setup and Results}
\label{sec:results}

We have tested our proposed technique on our Flickr8k dataset's subset, our own generated 'man' dataset, and the 'dog' dataset generated by previous year's group in the same course. Table \ref{tab:comparison} shows the BLEU performance of our model. It can be seen that our model is able to produce substantial BLEU score. We outperform the previous year's group by almost a double. Their BLEU score was 0.4 and we have managed to achieve a better BLEU score of 0.83. This can be further increased by the use of the discussed Urdu grammar correction techniques in the following section. The BLEU score on 'man' dataset is better because very specific and similar context based images were used and translations were also done very carefully. Another advantage is that in Urdu language much of words in sentences repeats frequently.

\begin{table}[]
\centering
\caption{Table showing the BLEU performance of our technique on the different datasets. Only the best results have been reported over here.}
\label{tab:comparison}
\begin{tabular}{|c|c|l|c|c|}
\cline{1-2} \cline{4-5}
\multicolumn{2}{|c|}{\textbf{\begin{tabular}[c]{@{}c@{}}Our Dataset\\ (Man-Urdu)\end{tabular}}} &  & \multicolumn{2}{c|}{\textbf{\begin{tabular}[c]{@{}c@{}}Other\\ Datasets\end{tabular}}} \\ \cline{1-2} \cline{4-5} 
\textbf{Models} & \textbf{BLEU} &  & \textbf{Dataset} & \textbf{BLEU} \\ \cline{1-2} \cline{4-5} 
Resnet-101-V2 & 0.83 &  & Flickr-English & 0.68 \\ \cline{1-2} \cline{4-5} 
Inception-V3 & 0.81 &  & Man-English & 0.66 \\ \cline{1-2} \cline{4-5} 
Xception & 0.8 &  & Dog-Urdu \cite{Kashif2019Caption} & 0.68 \\ \cline{1-2} \cline{4-5} 
\end{tabular}
\end{table}

\begin{figure*}[!ht]
  \centering
  \includegraphics[width=0.80\linewidth]{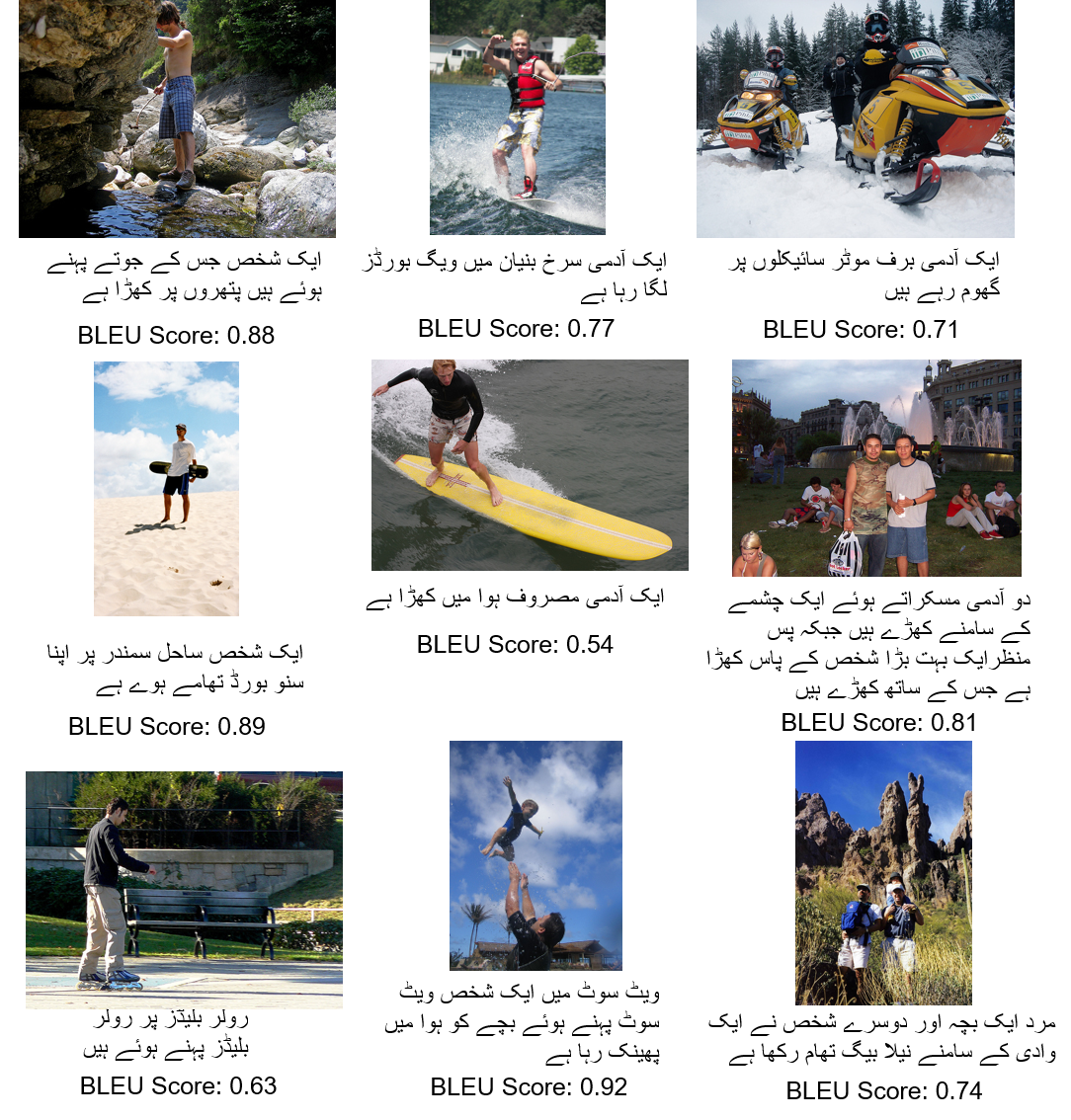}
  \caption{Predicted captions and BLEU scores of our model on 9 random images from validation data. It can be seen that our model performs sufficiently well on validation data.}
  \label{fig:result_collage}
\end{figure*}

The results of our model's attention mechanism are given in Figure \ref{fig:soft} and some predictions of our model on images from the validation dataset are shown in the figure \ref{fig:result_collage}. It can be seen that the model is able to perform sufficiently well in predicting the Urdu captions. A few wrong predictions can also be seen. 
All our codes can be found at this link \cite{mygit}.

\section{Discussion}
\label{sec:discussion}

The major focus of previous research has been on the western languages. Due to this, there are scarce resources of eastern languages like Urdu. This has been a huge hurdle in our work as we had to create the dataset ourselves. Urdu is comparatively complex as its morphology and syntax structure is a combination of Persian, Sanskrit, English, Turkish and Arabic \cite{adeeba2011experiences}.

Defining the word boundaries is a difficult task for Urdu owing to difference in joiner and non-joiners. Even if we are able to properly define the word boundaries, the compound words like "yahan-wahan", "torr-marror", "koh-paima" are needed to be treated as a single word. This is a major hurdle we faced as there is no complete dataset than can identify these words and hence help tokenizing the data. There are different techniques proposed like that of minimum edit distance that can be used to detect these but there are limitations due to the presence of outliers. The spaces must be ignored in case of tokenization of Urdu \cite{daud2017urdu}, in the following cases:

\begin{itemize}
    \item Compound words like "wazeer azam" etc
    \item Reduplication like "subah subah" etc
    \item Affixation like "sarmaya kaari" etc
    \item Proper nouns like "inaam ilahi" etc
    \item English Words
    \item Abbreviations and acronyms
\end{itemize}

The model is not able to detect the nouns clearly. It is not able to detect the difference between "karta hai" and "karti hai". This is a major problem faced while predicting using our trained model. Such problems are not faced in English.

Stemming is major part of NLP. The objective of stemming is to standardize words by reducing a word into its origin or root \cite{riaz2007challenges}. We need to stem the data and keep the morphemes. Morpheme is the unit of language that reflects a meaningful form of the word. These morphemes can help us in identifying the object and the action being performed inside the image very clearly. The predicted sentences can then be corrected grammar-wise. This will produce much better results than by using simple tokenization and no other pre-processing.

\begin{figure}[!t]
  \centering
  \includegraphics[width=1.0\linewidth]{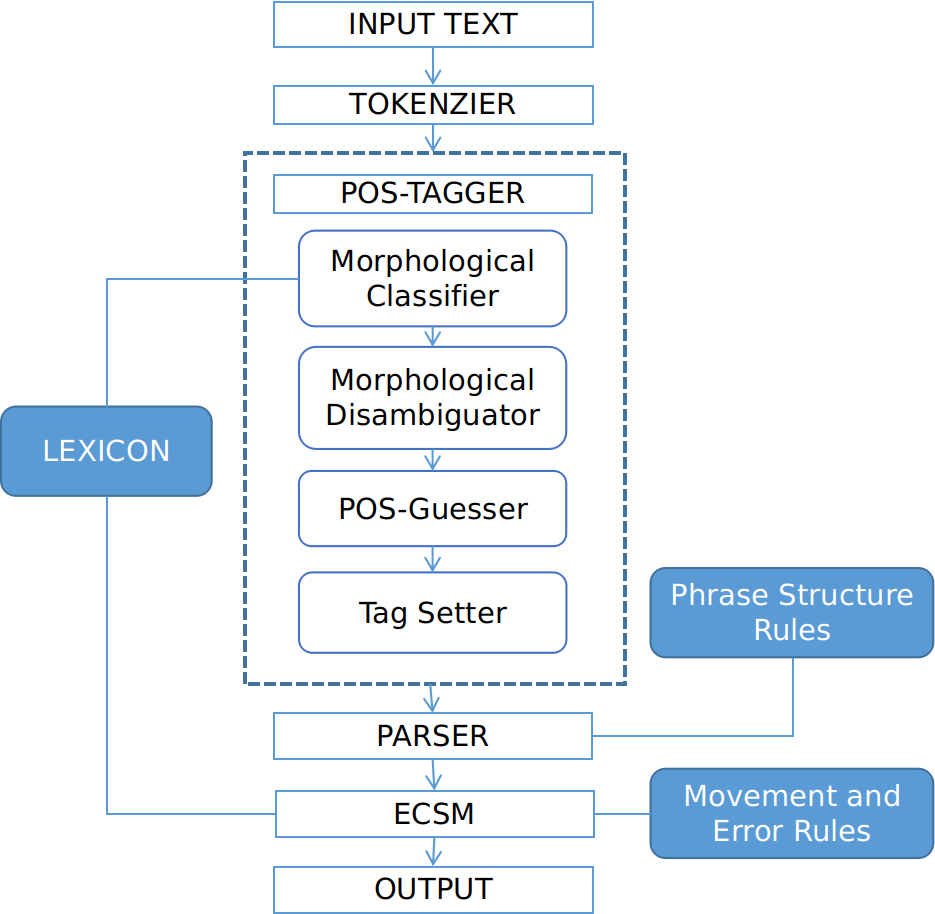}
  \caption{Figure showing the parsing algorithm proposed in \cite{kabir2002two}, where ECSM stands for "Error Checking and Suggestion Module".}
  \label{fig:parser}
\end{figure}

There are three types of techniques that can be used for Urdu language processing, namely: rule-based, statistical, and hybrid. As there are exceptions to every rule, so hybrid approach is the most suitable approach. Part-of-speech tagging can help us in finding the correct grammar and hence the model can learn to generate grammar-wise better captions. The unavailability of such a model for Urdu is a hurdle in our task.

Bhatt et al. \cite{bhatt2009multi} proposed a multi-representational and multi-layered tree-bank for Urdu. Their model proposes the dependency and phrase structure of the input sentence. Their model might have been a great help to our project but they have not made their code public and also they have performed the task for Roman Urdu while we are working on Nastaliq Urdu.

Kabir et al. \cite{kabir2002two} propose a two-pass parsing algorithm. In the first stage the model tries to predict the POS tagging of the input sentence. If it finds the input sentence to be inconsistent with the rules of Urdu grammar, then it suggests changes to the input sentences and after applying those changes the sentence is again passed to the parser. This suggestion of changes and again parsing is the second stage of the parsing algorithm. Along with grammatical mistakes it also looks for structural mistakes. The problem of unavailability of code was faced here. Their proposed work flow is shown in Figure \ref{fig:parser}.

\begin{figure}[!t]
  \centering
  \includegraphics[width=0.80\linewidth]{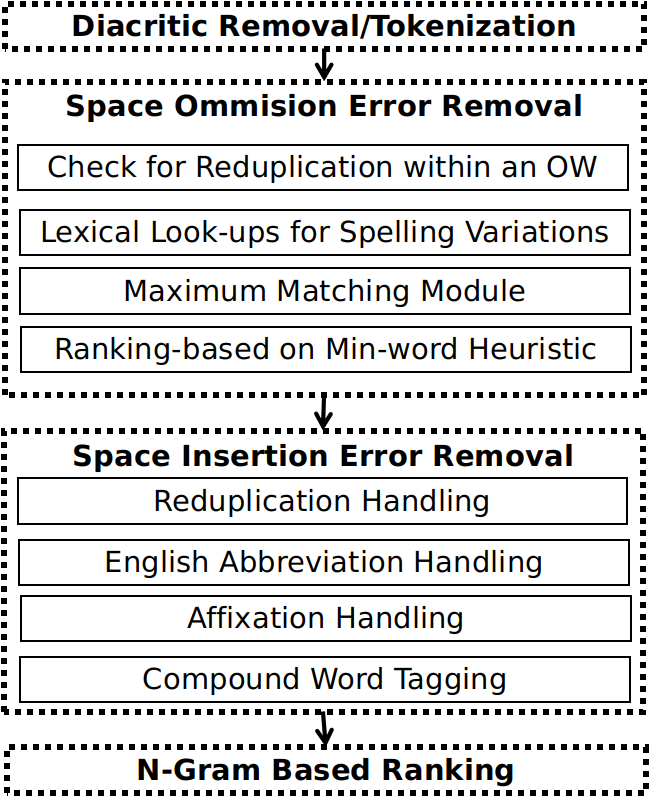}
  \caption{Figure showing the segmentation scheme proposed in \cite{durrani2010urdu}.}
  \label{fig:segment}
\end{figure}

Durrani et al. \cite{durrani2010urdu} propose a good workflow of Urdu language word segmentation. Figure \ref{fig:segment} shows the sequence that should be followed in order to pre-process the Urdu data to use it for any NLP task.

\section{Conclusions and Future Work}
\label{sec:conclusion}
We have prepared a dataset for Urdu language caption generation consisting of 700 images. Using this dataset we have proposed a deep learning algorithm that produces admirable results for similar unseen images. We show that our algorithm performs better than the previous year's group (that worked on the same problem) by almost a double margin. We have also provided a discussion on grammar correction of the proposed captions. Future work include devising a technique through which we can back-propagate the loss back from the grammar testing model to the GRU.

\small
\bibliographystyle{ieee}
\balance
\bibliography{report.bib}

\begin{thebibliography}{10}\itemsep=-1pt

\bibitem{adeeba2011experiences}
F.~Adeeba and S.~Hussain.
\newblock Experiences in building urdu wordnet.
\newblock In {\em Proceedings of the 9th workshop on Asian language resources},
  pages 31--35, 2011.

\bibitem{Kashif2019Caption}
{Afzal, Kashif and Raza, Kashif}.
\newblock {Urdu Language Caption Generation for Dog Dataset}.
\newblock Spring 2019 - Deep Learning Course's Semester Project at Information
  Technology University (ITU), Lahore, Pakistan, 2019.

\bibitem{bahdanau2014neural}
D.~Bahdanau, K.~Cho, and Y.~Bengio.
\newblock Neural machine translation by jointly learning to align and
  translate.
\newblock {\em arXiv preprint arXiv:1409.0473}, 2014.

\bibitem{bhatt2009multi}
R.~Bhatt, B.~Narasimhan, M.~Palmer, O.~Rambow, D.~M. Sharma, and F.~Xia.
\newblock A multi-representational and multi-layered treebank for hindi/urdu.
\newblock In {\em Proceedings of the Third Linguistic Annotation Workshop (LAW
  III)}, pages 186--189, 2009.

\bibitem{bybaby}
S.~by~Saheel.
\newblock Baby talk: Understanding and generating image descriptions.

\bibitem{cho2014learning}
K.~Cho, B.~Van~Merri{\"e}nboer, C.~Gulcehre, D.~Bahdanau, F.~Bougares,
  H.~Schwenk, and Y.~Bengio.
\newblock Learning phrase representations using rnn encoder-decoder for
  statistical machine translation.
\newblock {\em arXiv preprint arXiv:1406.1078}, 2014.

\bibitem{chung2014empirical}
J.~Chung, C.~Gulcehre, K.~Cho, and Y.~Bengio.
\newblock Empirical evaluation of gated recurrent neural networks on sequence
  modeling.
\newblock {\em arXiv preprint arXiv:1412.3555}, 2014.

\bibitem{daud2017urdu}
A.~Daud, W.~Khan, and D.~Che.
\newblock Urdu language processing: a survey.
\newblock {\em Artificial Intelligence Review}, 47(3):279--311, 2017.

\bibitem{de2006generating}
M.-C. De~Marneffe, B.~MacCartney, C.~D. Manning, et~al.
\newblock Generating typed dependency parses from phrase structure parses.
\newblock In {\em Lrec}, volume~6, pages 449--454, 2006.

\bibitem{durrani2010urdu}
N.~Durrani and S.~Hussain.
\newblock Urdu word segmentation.
\newblock In {\em Human Language Technologies: The 2010 Annual Conference of
  the North American Chapter of the Association for Computational Linguistics},
  pages 528--536, 2010.

\bibitem{fang2015captions}
H.~Fang, S.~Gupta, F.~Iandola, R.~K. Srivastava, L.~Deng, P.~Doll{\'a}r,
  J.~Gao, X.~He, M.~Mitchell, J.~C. Platt, et~al.
\newblock From captions to visual concepts and back.
\newblock In {\em Proceedings of the IEEE conference on computer vision and
  pattern recognition}, pages 1473--1482, 2015.

\bibitem{farhadi2010every}
A.~Farhadi, M.~Hejrati, M.~A. Sadeghi, P.~Young, C.~Rashtchian, J.~Hockenmaier,
  and D.~Forsyth.
\newblock Every picture tells a story: Generating sentences from images.
\newblock In {\em European conference on computer vision}, pages 15--29.
  Springer, 2010.

\bibitem{he2016deep}
K.~He, X.~Zhang, S.~Ren, and J.~Sun.
\newblock Deep residual learning for image recognition.
\newblock In {\em Proceedings of the IEEE conference on computer vision and
  pattern recognition}, pages 770--778, 2016.

\bibitem{hinton2012neural}
G.~Hinton, N.~Srivastava, and K.~Swersky.
\newblock Neural networks for machine learning lecture 6a overview of
  mini-batch gradient descent.
\newblock {\em Cited on}, 14(8), 2012.

\bibitem{hodosh2013framing}
M.~Hodosh, P.~Young, and J.~Hockenmaier.
\newblock Framing image description as a ranking task: Data, models and
  evaluation metrics.
\newblock {\em Journal of Artificial Intelligence Research}, 47:853--899, 2013.

\bibitem{huang2017densely}
G.~Huang, Z.~Liu, L.~Van Der~Maaten, and K.~Q. Weinberger.
\newblock Densely connected convolutional networks.
\newblock In {\em Proceedings of the IEEE conference on computer vision and
  pattern recognition}, pages 4700--4708, 2017.

\bibitem{mygit}
{Inaam Ilahi}.
\newblock {Urdu Language Caption Generation, Deep Learning Project}.
\newblock Link:
  \url{https://github.com/Inaam1995/DeepLearning_Project_Spring2020}, 2020.
\newblock Online; accessed 6 February 2021.

\bibitem{jozefowicz2016exploring}
R.~Jozefowicz, O.~Vinyals, M.~Schuster, N.~Shazeer, and Y.~Wu.
\newblock Exploring the limits of language modeling.
\newblock {\em arXiv preprint arXiv:1602.02410}, 2016.

\bibitem{kabir2002two}
H.~Kabir, S.~Nayyer, J.~Zaman, and S.~Hussain.
\newblock Two pass parsing implementation for an urdu grammar checker.
\newblock In {\em Proceedings of IEEE international multi topic conference},
  pages 1--8, 2002.

\bibitem{karpathy2014deep}
A.~Karpathy, A.~Joulin, and L.~F. Fei-Fei.
\newblock Deep fragment embeddings for bidirectional image sentence mapping.
\newblock In {\em Advances in neural information processing systems}, pages
  1889--1897, 2014.

\bibitem{kingma2014adam}
D.~P. Kingma and J.~Ba.
\newblock Adam: A method for stochastic optimization.
\newblock {\em arXiv preprint arXiv:1412.6980}, 2014.

\bibitem{kiros2014multimodal}
R.~Kiros, R.~Salakhutdinov, and R.~Zemel.
\newblock Multimodal neural language models.
\newblock In {\em International conference on machine learning}, pages
  595--603, 2014.

\bibitem{kiros2014unifying}
R.~Kiros, R.~Salakhutdinov, and R.~S. Zemel.
\newblock Unifying visual-semantic embeddings with multimodal neural language
  models.
\newblock {\em arXiv preprint arXiv:1411.2539}, 2014.

\bibitem{kuznetsova2012collective}
P.~Kuznetsova, V.~Ordonez, A.~Berg, T.~Berg, and Y.~Choi.
\newblock Collective generation of natural image descriptions.
\newblock In {\em Proceedings of the 50th Annual Meeting of the Association for
  Computational Linguistics (Volume 1: Long Papers)}, pages 359--368, 2012.

\bibitem{li2011composing}
S.~Li, G.~Kulkarni, T.~Berg, A.~Berg, and Y.~Choi.
\newblock Composing simple image descriptions using web-scale n-grams.
\newblock In {\em Proceedings of the Fifteenth Conference on Computational
  Natural Language Learning}, pages 220--228, 2011.

\bibitem{lin2014microsoft}
T.-Y. Lin, M.~Maire, S.~Belongie, J.~Hays, P.~Perona, D.~Ramanan,
  P.~Doll{\'a}r, and C.~L. Zitnick.
\newblock Microsoft coco: Common objects in context.
\newblock In {\em European conference on computer vision}, pages 740--755.
  Springer, 2014.

\bibitem{mao2014deep}
J.~Mao, W.~Xu, Y.~Yang, J.~Wang, Z.~Huang, and A.~Yuille.
\newblock Deep captioning with multimodal recurrent neural networks (m-rnn).
\newblock {\em arXiv preprint arXiv:1412.6632}, 2014.

\bibitem{riaz2007challenges}
K.~Riaz.
\newblock Challenges in urdu stemming (a progress report).
\newblock In {\em BCS IRSG Symposium: Future Directions in Information Access
  2007}, pages 1--6, 2007.

\bibitem{sutskever2013importance}
I.~Sutskever, J.~Martens, G.~Dahl, and G.~Hinton.
\newblock On the importance of initialization and momentum in deep learning.
\newblock In {\em International conference on machine learning}, pages
  1139--1147, 2013.

\bibitem{szegedy2016rethinking}
C.~Szegedy, V.~Vanhoucke, S.~Ioffe, J.~Shlens, and Z.~Wojna.
\newblock Rethinking the inception architecture for computer vision.
\newblock In {\em Proceedings of the IEEE conference on computer vision and
  pattern recognition}, pages 2818--2826, 2016.

\bibitem{vinyals2015show}
O.~Vinyals, A.~Toshev, S.~Bengio, and D.~Erhan.
\newblock Show and tell: A neural image caption generator.
\newblock In {\em Proceedings of the IEEE conference on computer vision and
  pattern recognition}, pages 3156--3164, 2015.

\bibitem{xu2015show}
K.~Xu, J.~Ba, R.~Kiros, K.~Cho, A.~Courville, R.~Salakhudinov, R.~Zemel, and
  Y.~Bengio.
\newblock Show, attend and tell: Neural image caption generation with visual
  attention.
\newblock In {\em International conference on machine learning}, pages
  2048--2057, 2015.

\bibitem{flickr30k}
P.~Young, A.~Lai, M.~Hodosh, and J.~Hockenmaier.
\newblock From image descriptions to visual denotations: New similarity metrics
  for semantic inference over event descriptions.
\newblock {\em TACL}, 2:67--78, 2014.

\end{thebibliography}

\end{document}